\documentclass[runningheads]{llncs}
\usepackage[T1]{fontenc}

\usepackage{amsmath}
\usepackage{amsfonts}
\usepackage{graphicx}
\usepackage{anyfontsize}
\usepackage{xcolor}
\usepackage[
    colorlinks=true,
    pdfborder={0 0 0},
    linkcolor=red
]{hyperref}

\def\debug{1}

\ifnum \debug=1
\providecommand{\lnote}[1]{\textcolor{blue}{[LR: #1]}}
\providecommand{\mnote}[1]{\textcolor{violet}{[MB: #1]}}
\providecommand{\anote}[1]{\textcolor{orange}{[MA: #1]}}
\providecommand{\pnote}[1]{\textcolor{red}{[AP: #1]}}
\else
\providecommand{\lnote}[1]{}
\providecommand{\mnote}[1]{}
\providecommand{\anote}[1]{}
\providecommand{\pnote}[1]{}
\fi

\begin{document}

\title{CFTS-GAN: Continual Few-Shot Teacher Student for Generative Adversarial Networks}
\author{
Munsif Ali
\and
Leonardo Rossi
\and 
Massimo Bertozzi}

\authorrunning{M. Ali et al.}

\institute{Department of Engineering and Architecture, University of Parma, Italy
\email{\{munsif.ali, leonardo.rossi, massimo.bertozzi\}@unipr.it}}

\titlerunning{CTFS-GAN}
%

%
\maketitle

\begin{abstract}
Few-shot and continual learning face two well-known challenges in GANs: overfitting and catastrophic forgetting. Learning new tasks results in catastrophic forgetting in deep learning models. In the case of a few-shot setting, the model learns from a very limited number of samples (e.g. 10 samples), which can lead to overfitting and mode collapse.
So, this paper proposes a Continual Few-shot Teacher-Student technique for the generative adversarial network (CFTS-GAN) that considers both challenges together. Our CFTS-GAN uses an adapter module as a student to learn a new task without affecting the previous knowledge. To make the student model efficient in learning new tasks, the knowledge from a teacher model is distilled to the student. In addition, the Cross-Domain Correspondence (CDC) loss is used by both teacher and student to promote diversity and to avoid mode collapse. Moreover, an effective strategy of freezing the discriminator is also utilized for enhancing performance. Qualitative and quantitative results demonstrate more diverse image synthesis and produce qualitative samples comparatively good to very stronger state-of-the-art models.


\end{abstract}

\section{Introduction}\label{introduction}
Continual Learning (CL) and Few-shot (FS)  learning are two very important problems to take into account, which affect the applicability of deep learning systems in real cases \cite{wang2024comprehensive,wang2020generalizing}.
Continual learning makes the model capable of learning new tasks without affecting the previously learned tasks. 
However, learning a new task in deep learning models leads to a well-known problem named catastrophic forgetting~\cite{le2023mode,song2024overcoming}. 
In a few-shot setting, a model learns only from a few samples (e.g.~$\leq10$), which can result in memorizing the training set and causes overfitting and mode collapse~\cite{abdollahzadeh2023survey}.  
FS is also a considerable matter because of the unavailability of the full dataset due the data privacy concerns. Moreover,  FS is an important consideration for such applications where very limited samples are available such as healthcare, cyber security, etc. Furthermore, it is also essential due to huge resource consumption, training time, and low-power devices~\cite{wang2024comprehensive,wang2020generalizing}.
Individually, continual and few-shot learning are very well explored; conversely, considering both together have gained more popularity recently \cite{abdollahzadeh2023survey}.
So, we also take into consideration both together and present our contributions to the challenging tasks in this paper. 

In CL and FS there are two very distinct areas of application of the techniques: the ones that apply to discriminative models, such as classifiers, segmentation models, etc., and the ones that apply to generative models, such as GAN models. 
E.g. a survey article in \cite{tian2024survey} collected the works that considered both the CL and FS together for classification and segmentation. 
Various techniques exist that focus on both few-shot and continual learning to mitigate forgetting and overfitting in classification models. 
However, generative models still need more attention and consideration regarding continual and few-shot learning and have gained attention recently \cite{abdollahzadeh2023survey}.  Therefore, we consider the CL and FS which apply to generative models, especially to GANs.

The CL approaches are grouped into regularization \cite{ewc_g}, replay  \cite{li2024adaer}, or dynamics/expansion methods  \cite{cam_gan,pnn,den}.  Regularization approaches grant to overcome forgetting but produce blurred samples after learning many tasks~\cite{cam_gan}. In replay approaches, memory consideration limits the scalability of these approaches, and data privacy is also an important consideration where data privacy is concerned~\cite{gan_cl}.  Dynamic architectures add additional parameters for CL, do not need previous data samples, and also provide good results. However, designing such architecture needs careful attention due to increasing the number of parameters \cite{cam_gan}.
On the other hand, when limited samples are available for training, instead of learning the data distribution, the model memorizes the training samples and leads to overfitting and mode collapse. Earlier approaches for FS and limited data generation used transfer learning and fine-tuning for the target domain generation~\cite{abuduweili2021adaptive,pan2009survey,wang2018transferring}.  Another approach is data augmentation either for the data or feature level for gaining diversity in the target generation~\cite{wang2024patch}. However, these approaches are not a good choice in the case of very limited training samples \cite{israr2022customizing}. In the knowledge distillation, another model is used to make the deep learning model efficient considering the FS setting \cite{park2024pre}.  However, all these approaches do not consider both the CL and FS image generation.  Less work has been done considering both few-shot and continual learning together for GANs \cite{seo2023lfs}. 
Our work proposes a continual few-shot teacher-student model for GANs (CFTS-GAN) considering the FS and CL together.
Our method uses knowledge distillation, adding regularization terms for the few-shot image generation. 
Our CFTS-GAN model takes inspiration from CAM-GAN~\cite{cam_gan} for continual learning and from \cite{ojha2021few} for few-shot learning. 
CAM-GAN injects adapter modules on the top of a generator model~\cite{mescheder2018training} for continual image generation.
It trains only the adapters when a new task is available, preserving the ability to generate images for the previous tasks.   
The CAM-GAN consists of a simple generator architecture compared to StyleGAN2~\cite{karras2020analyzing}, whereas StyleGAN2 has more control over the image generation due to latent space manipulation and adding noise in every stage. 
We extend the CAM-GAN training with the teacher-student architecture to improve its performance and with the CDC loss to preserve diversity and avoid mode collapse.
Our final architecture consists of three generators: a source, a teacher, and a student.
Starting from a source generator, previously trained on a large dataset, we train CFTS-GAN in two stages.
In the first stage, the teacher model is trained on the current task, preserving the generator diversity with the help of the CDC loss~\cite{ojha2021few} between the source model and itself.
In the second stage, when the student model is trained on the current task, the student takes advantage of the teacher, squeezing teacher knowledge inside the adapters and, at the same time, preserving generator diversity using CDC loss between the source model and itself. 
By applying CDC loss to both, we decrease the probability that the generators lose the ability to generate different images.
CFTS-GAN also utilizes a simple technique of freezing the student's discriminator to obtain much better results \cite{mo2020freeze}. 
The quantitative results demonstrate that our approach gains more diversity and obtains comparatively quality samples compared with stronger models \cite{seo2023lfs}, \cite{ojha2021few}, \cite{xiao2022few}, and \cite{zhao2022closer} which are derived from a more advanced architecture \cite{karras2020analyzing}.


The main contributions of this paper are summarized in the following points.

\begin{itemize}
  \item We propose a teacher-student model for continual few-shot image synthesis, to condense the knowledge of a generator into the adapters of a CAM-GAN model.
  \item To preserve the diversity of the image generated and prevent mode collapse due to memorization of few available examples, we employed the Cross-Domain Correspondence (CDC) loss \cite{ojha2021few} in both the teacher and student models training, introducing a source model pre-trained on a large dataset.
  \item To further refine the model performance, a simple strategy of freezing the discriminator in the student training is also used \cite{mo2020freeze}.
  \item To evaluate image quality and diversity, the performance of the CFTS-GAN is analyzed on different few-shot datasets using FID and B-LPIPS \cite{seo2023lfs} metrics. 
\end{itemize}


\section{Literature Review}\label{literature}

\noindent\textbf{Continual Learning}.\label{cl-gan}
Generative models have been recently analyzed to generate data continually and get rid of overriding new information on the older ones. A well-known approach in~\cite{ewc} is proposed for discriminative models where an additional loss term is added to retain the previously learned distribution. The idea of~\cite{ewc} is utilized and implemented in GAN~\cite{ewc_g} where an additional loss term is added only to the generator of the conditional GAN to prevent previously learned distribution from drifting and avoid forgetting. However, after too many tasks the model saturates and provides an unrealistic generation of images.  Another notable approach is memory replay GAN (MeRGAN)~\cite{chenshen2018memory} in which a previous sample is recalled during training for the current task. The previous data sample is generated using the generator which solves the issue but still provides unrealistic samples after many tasks. Moreover, rehearsal-based approaches are limited to label conditional generation~\cite{zhai2019lifelong}. To implement both label and image conditional GAN, the authors in \cite{zhai2019lifelong} proposed an approach using knowledge distillation for a lifelong generation. However, the proposed lifelong model is shared across all tasks so the previous task generation quality is degraded as new tasks arrive.  
In parameter isolation-based approaches, the parameters of the previous tasks are kept unchanged while learning new tasks. One way of parameter isolation is to expand the network and each task has its own parameters~\cite{kumar2021bayesian,yan2021dynamically}. Another way is to use the same architecture but allocate distinct parameters for each task~\cite{mallya2018piggyback,rajasegaran2019random}. For continual classification, a model is designed using universal and parameter vectors for learning shared and task-specific domains, respectively~\cite{rebuffi2018efficient}. In~\cite{pnn}, add additional layers for learning a new task and use previously learned knowledge, and fine-tune for better convergence.
CAM-GAN authors~\cite{cam_gan} proposed a continual learning approach that is based on adding adapter modules for learning the upcoming tasks without affecting the previously learned distributions. 
Taking inspiration from them, we extended their model to work on a more challenging scenario of CL and FS, together.
We also use their adapters concept on a Teacher-Student setting to enhance generation quality and CDC loss to preserve diversity of generated images.

\noindent\textbf{Few-Shot Learning}.
Many methods exist for the few-shot GANs \cite{duan2024weditgan,shi2023autoinfo,sushko2023smoothness}. 
A simple fine-tuning FreezD method based on freezing the discriminator layers to achieve the few-shot image generation is presented in \cite{mo2020freeze}. Transfer learning is also considered for the few-shot image synthesis in \cite{wang2018transferring}. BSA \cite{noguchi2019image} adds a small number of parameters to the source model based on the statistics of the feature map. CDC \cite{ojha2021few} preserves the diversity of the images by applying cross-domain correspondence between the source and target images. Other variants of CDC are also presented in \cite{xiao2022few} and \cite{zhao2022closer}. Similar to CDC the DCL \cite{zhao2022closer} uses the mutual information between the source and target domain. 
These approaches consider only few-shot learning while in our case we take into consideration both continual and few-shot learning.

\noindent\textbf{Continual Few-Shot Learning}.
For discriminative models, a hyper-transformer is used for few-shot lifelong learning classification in \cite{vladymyrov2022few}. The hyper-transformer generates the weights for learning the new task. The authors in \cite{tao2020few} proposed another approach for the few-shot class incremental classification using a neural gas network. An expansion-based idea is presented in \cite{zhou2022forward} for few-shot discrimination which tries to make the same sample closer and increase the space between different samples for few-shot settings. Conversely, generative models in terms of both continual and few-shot settings are not widely explored. According to the authors, the LFS-GAN~\cite{seo2023lfs} is the first approach that considers both setups together. LFS-GAN appends additional parameters for learning new tasks. The distance between generated fake samples and input noise is maximized to produce diverse images. A continual few-shot image translation is proposed in~\cite{chen2022few}. This method introduces new scale and shift parameters and modulates the older parameters to learn the newly introduced scale and shift parameters. 
We proposed different training approaches for continual few-shot learning utilizing the architecture of \cite{cam_gan} introducing teacher-student architecture. Moreover, we also applied CDC loss in both training stages to obtain better diversity. 

\section{Method}\label{method}
In this section, we describe our approach for the continual few-shot image generation which is divided into several subsections. The sections \ref{cfs} and \ref{cdc_loss} give details of the preliminaries of continual and few-shot learning and CDC loss. Section \ref{teacher_student} describes the teacher-student model and their objectives. 

\begin{figure}[t!]
    \centering
    \includegraphics[width=0.7\columnwidth]{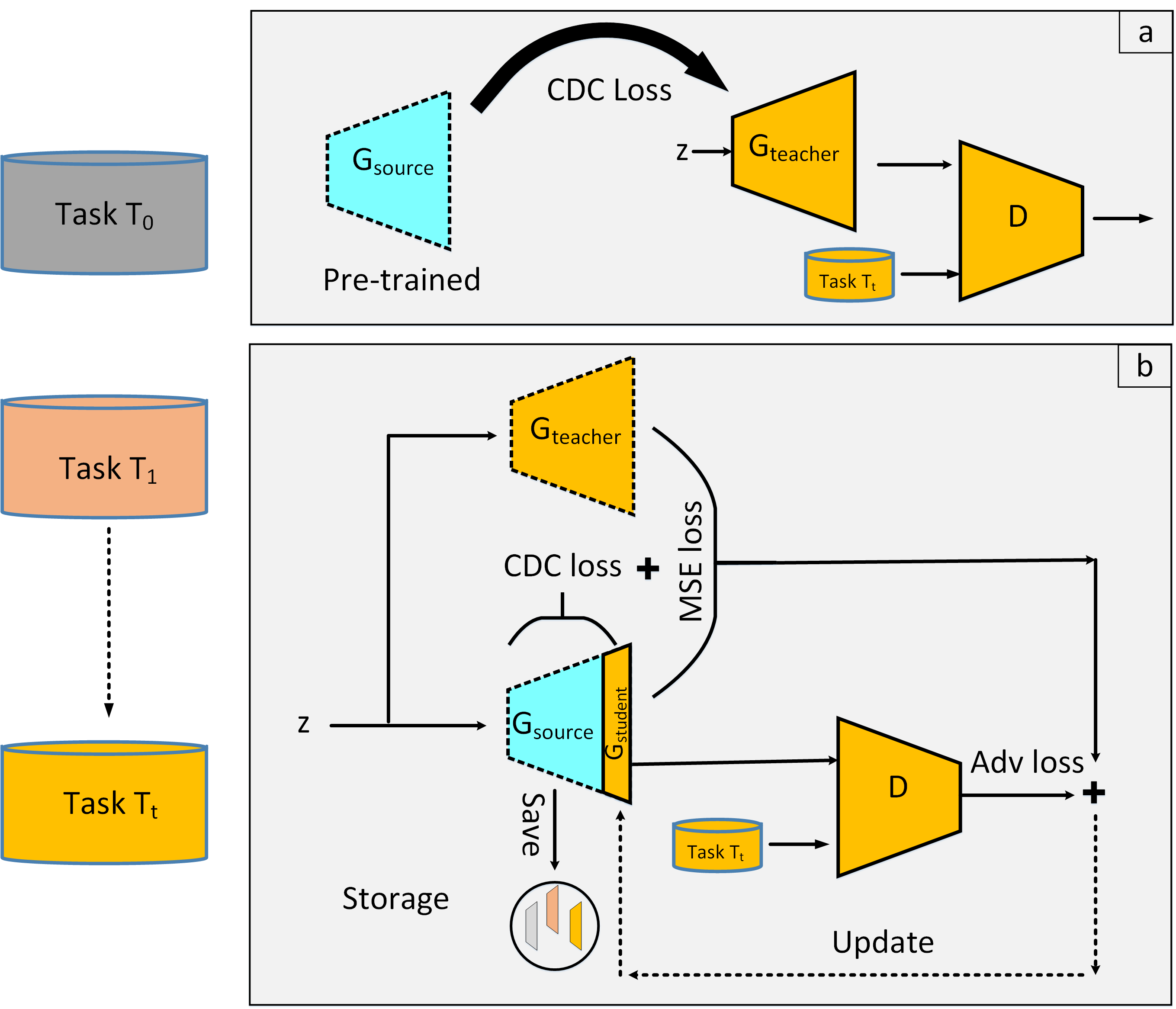}
    \caption{The teacher-student model for continual learning GAN. The dashed lines show the frozen model. a) shows the teacher model training. b) demonstrates training of the student model.}
    \label{fig:model.}
\end{figure}

\subsection{Continual Few-Shot GAN}\label{cfs}
For continual few-shot image synthesis, a set of multiple tasks $T = \{T_0, .. , T_t\}$ is considered (Figure~\ref{fig:model.}).
Each task $T_i$ consists of a dataset with few samples ${x_0, x_2, ..... x_n}$, where $n$ is the total number of samples in the training set.
If $n$ is low (i.e. $ n \leq 10$), we are in the case of few-shot learning. 
As in CL setting, when the model is trained on a new task $T_i$, previous training data are no longer available and then can not be used for this training step.

GANs are usually composed by two models: a discriminator D and generator G. The objective of D is to discern real data from synthetic samples generated by G. 
While the objective of G is to generate samples that look real to fool D.
The generator learns parameters $\theta$ from a probability distribution $p_{data}(x)$ for a given dataset to produce synthetic data. 
These fake images are generated from a random Gaussian distribution $p(z)$.  
Both D and G are trained together in an adversarial manner to achieve the following objective \cite{goodfellow2020generative}

\begin{equation}\label{adversarial}
\min_{G}\max_{D}\mathbb{E}_{x\sim p_{\text{data}}(x)}[\log{D(x)}] + \mathbb{E}_{z\sim p_{\text{z}}(z)}[1 - \log{D(G(z))} ] + R(\theta_d)
\end{equation}

\noindent where $p_{data}(x)$ is the distribution of real data and $p_z(z)$ is a random Gaussian distribution noise, $R(.)$ is a regularization term \cite {mescheder2018training}.

For each task $T_i$, our training is composed by two stages. 
In the first stage, a teacher model is trained on the FS dataset for task $T_i$. In the second stage, to exploit the knowledge distillation \cite{aguinaldo2019compressing} using the teacher, the student model is trained. 
To enhance the diversity of the generated examples and avoid mode collapse, the Cross Domain Consistency (CDC) loss is employed for both training, teacher, and student, in addition to adversarial loss, as explained in the following paragraphs.

In the case of FS, due to limited samples, the discriminator part of the GAN also overfits and affects the generator training. So, freezing the discriminator is one of the strategies to avoid the overfitting of it \cite{mo2020freeze}.

\subsection{Cross Domain Consistency Loss}\label{cdc_loss}

Considering that few-shot learning leads to mode collapse in GANs, data augmentation is one possible solution for diverse image generation but it works better when the training samples are in a much larger number \cite{israr2022customizing}. 
Since we are in a much more restrictive situation, to preserve the source diversity and avoid mode collapse, we took inspiration from the \cite{ojha2021few}.
During the training process of the generator (teacher or student, depending if we are in the first or second stage) on the dataset for the task $T_i$, another frozen generator, called source, is incorporated into the training procedure.

As previously mentioned, we consider two generators in this scenario: $G_{source}$, trained on large dataset (in our case, trained on the FFHQ dataset), and $G_{target}$, which is trained using a few-shot learning method on dataset for the task $T_i$.
At each training step $i$, two noise vectors $z_j$ and $z_k$ are sampled and passed to both the $G_{source}$ and $G_{target}$. 
The probability distribution between different noise vectors for the source and target is given as

\begin{equation}\label{cdc}
\begin{aligned}
& y_{source}=\operatorname{Softmax}\left(\left\{\operatorname{sim}\left(G_{source}\left(z_j\right), G_{source}\left(z_k\right)\right)\right\}_{\forall j \neq k}\right) \\
& y_{target}=\operatorname{Softmax}\left(\left\{\operatorname{sim}\left(G_{target}\left(z_j\right), G_{target}\left(z_k\right)\right)\right\}_{\forall j \neq k}\right)
\end{aligned}
\end{equation}

\noindent where $y_{source}$ and  $y_{target}$ are the probability of the source and target, $sim$ denotes the cosine similarity, and $G_{source}(z)$ and $G_{target}(z)$ are the output layer for the source and target generators.
We use the KL divergence to encourage the target generator to have a similar distribution of the source generator.
This loss is called CDC loss and is given as:

\begin{equation}
\mathcal{L}_{\operatorname{cdc}}\left(G_{source}, G_{target}\right)= KL\left(y_{target} \| y_{source}\right).
\end{equation}

In our case, our target generator $G_{target}$ will be the teacher or the student, depending if we are in the first or second stage.
In the following sections, we will use the notation $G_{teacher}$ and $G_{student}$ instead of the target $G_{target}$ for the corresponding teacher and student training.

\subsection{Teacher Student Model}\label{teacher_student}

Our generator model, based on CAM-GAN \cite{cam_gan}, injects adapter modules on the top of GP-GAN \cite{mescheder2018training} for continual learning. 
The CAM-GAN consists of global weights $\theta_{global}$ and adapter weights $\theta_{adapter}$.
The latter are used for learning the upcoming tasks while freezing the global weights, so as not to affect the previous tasks generation.

Our continual few-shot image generation training is composed of two stages, and takes into account three generators: source, teacher, and student.
All of them have the same architecture but each one is used in a particular setting and different purpose.
At the beginning, the source is trained on a very large dataset.
As show in Figure \ref{fig:model.}(a), in the first stage the teacher model is cloned from the source model and trained all its weights $\theta_{teacher}$.
Because also the teacher is trained in a few-shot setting, to enforce the diversity, we support its training with the CDC loss \cite{ojha2021few}, in addition to its adversarial loss.

As show in Figure \ref{fig:model.}(b), in the second stage the student model is trained.
To achieve the continual image generation objective, we considered the student model weights $\theta_{student}$ as the combination of global weights $\theta_{global}$ and adapter weights $\theta_{adapter}$, initially cloned from the source. 
To learn the current few-shot task $T_i$, only the adapters' weights are trained, while freezing the global weights.  
To make the student more effective in learning the current few-shot task $T_i$, it takes advantage from the source using CDC loss \cite{ojha2021few}, and from the teacher model, previously trained with the same dataset, using a loss of knowledge distillation~\cite{aguinaldo2019compressing}.

\noindent\textbf{Teacher objective}.\label{teacher_loss}
The teacher learning is shown in Figure \ref{fig:model.}(a). 
The teacher model objective uses the adversarial loss as given in the equation \ref{adversarial}.  For diverse image generation, the CDC loss is used along with the adversarial loss. So, the objective of the teacher is the sum of the adversarial loss and CDC loss for the teacher $G_{teacher}$ and is given by
\begin{equation}
\mathcal{L}_{teacher} = \mathcal{L}_{adv} + w_t \mathcal{L}_{\operatorname{cdc}}\left(G_{source}, G_{teacher}\right),
\end{equation}

\noindent where $w_t$ is a scalar weight factor for the CDC loss. 
Unlike starting from scratch, the teacher model's weights are initialized with the source parameters.

\noindent\textbf{Student Objective}.\label{student_loss}
The training process of the student model is depicted in Figure \ref{fig:model.}(b). To train the student model $G_{student}$, the knowledge from the teacher model $G_{teacher}$ is transferred to the student. To transfer the knowledge from the teacher model to the student we utilized the Mean Square Error (MSE) loss~\cite{aguinaldo2019compressing}, termed as $\mathcal{L}_{kd}$, applied to the output. 
This loss minimizes the loss between the teacher and student and its objective is given by

\begin{equation}\label{MSE}
\mathcal{L}_{kd} = \frac{1}{N}\sum_{i=1}^{N}||G_{teacher} (z)- G_{{student}} (z)||^2.
\end{equation}

Moreover, the objective of the student adapter modules is to fool the discriminator, minimizing the standard GAN loss

\begin{equation}
 \mathcal{L}_{adv} = \frac{1}{N}\sum_{i=1}^{N} \log (1-D(G_{{student}}(z))),
\end{equation}
where D means discriminator network of the GANs. Lastly, the $\mathcal{L}_{kd}$ and the standard GAN loss are combined with the CDC loss to obtain more diversity. The final objective of the student is given by 

\begin{equation}
\mathcal{L}_{student} = \mathcal{L}_{adv} + \alpha \mathcal{L}_{kd} + w_s \mathcal{L}_{\operatorname{cdc}}
\end{equation}
where $\alpha$ is the weight of the  loss $\mathcal{L}_{kd}$ and $w_s$ is the weight of the CDC loss for the student.

\section{Experiments}\label{results}
\textbf{Datasets.}
This section provides details of the performed experiments and shows the qualitative and quantitative results of the CFTS-GAN for continual few-shot image synthesis. The datasets taken into consideration for the experiments are sketches \cite{ojha2021few}, female \cite{karras2017progressive}, sunglasses \cite{ojha2021few}, male \cite{karras2017progressive}, and babies \cite{ojha2021few}. These datasets are used as the target datasets for evaluating the efficacy of our model for the few-shot continual generation of images. Each of these datasets consists of 10 samples. While the source model is pre-trained on a large FFHQ dataset~\cite{karras2019style}.

\noindent\textbf{Evaluation metrics.}
The state-of-the-art CDC \cite{ojha2021few}, RSSA \cite{xiao2022few}, DCL \cite{zhao2022closer},  CAM-GAN \cite{cam_gan} and LFS-GAN  \cite{seo2023lfs} approaches are considered for comparison. The CAM-GAN model appends adapter modules on the top of \cite{mescheder2018training} for continual image generation. The LFS-GAN adds more weights for the subsequent tasks using few-shot learning. Where LFS-GAN derives its model from a very strong model StyleGAN2 \cite{karras2020analyzing}. Moreover, it also utilizes the patch discriminator \cite{ojha2021few} to improve the target generation. While others compared  state-of-the-art models are also based on a stronger model \cite{karras2020analyzing}. We evaluated our CFTS-GAN teacher-student model for the few-shot continual learning, which produces diverse images and has a quality near to the stronger LFS-GAN model. The Fréchet inception distance (FID) \cite{heusel2017gans} and B-LPIPS \cite{seo2023lfs} are used as quantitative metrics. The FID score provides how much the synthesis images are close to the real images. The lower FID means that the generator produces samples closer to the real images. While the B-LPIPS shows the diversity of the images, the higher score shows a more diverse image generation.

\noindent\textbf{Training.} 
The source model is pre-trained on the source task using a large dataset using the FFHQ dataset. The FFHQ contains 70k high-resolution image samples. The input image dimension fed into the model is \(256 \times 256\). The other datasets mentioned above are considered as subsequent tasks. Therefore, the teacher and student model are trained to generate new tasks continually. For the training, only the data from the current task is available while the previous data is not available.  The teacher model is trained on a few shot images. The knowledge is then transferred from the teacher to the student model using knowledge distillation while also maintaining the source diversity. So the teacher-student model correspondingly produces quality and diverse samples without affecting the previous knowledge. 

\subsection{Qualitative Results}
\begin{figure*}[t!]
\centerline{\parbox{4.8 in}{~\hfill \parbox{10em}{\centering CAM-GAN} \hfill~\hfill \parbox{10em}{\centering LFS-GAN} \hfill~\hfill \parbox{10em}{\centering CTFS-GAN} \hfill~}}
  \centerline{\includegraphics[width= 4.8 in]{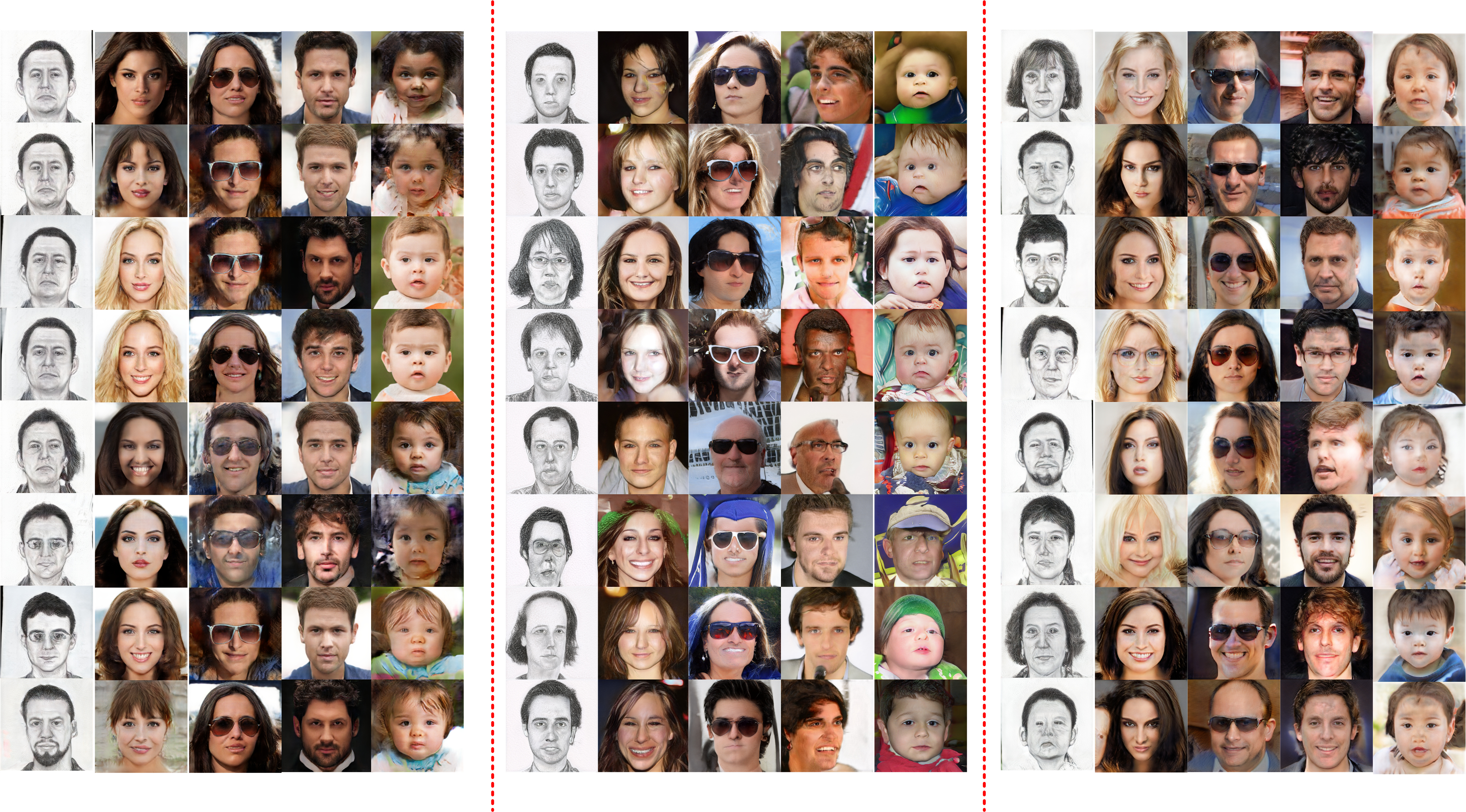}}
  \caption{Qualitative results: Generated samples. In each group, the first column is for sketches (Task 1), and the second, third, fourth, and fifth are for females (Task 2), sunglasses (Task 3), males (Task 4), and babies (Task 5).}\label{q-images}
\end{figure*}

The qualitative results are shown in Figure \ref{q-images}. Our model is able to generate different, diverse, and comparatively quality samples of images continually. The model produces the current data samples without affecting previously learned samples. A few generated samples from each task are presented in the paper.   

\subsection{Quantitative Results} 
The comparison in terms of FID and B-LPIPS scores with the state-of-the-art models is shown in Table \ref{fid-blpips}. The bold value represents the best results while the underlined values represent the second-best results. The CFTS-GAN represents the best performance than CAM-GAN, CDC, RSSA, and DCL in terms of both B-LPIPS and FID, maintaining the same number of weights and same architecture on evaluation as CAM-GAN.
Except for the FID of babies, the CDC is the second best. While compared with LFS-GAN the teacher-student model also gives the best performance in terms of B-LPIPS. In terms of FID, the LFS-GAN \cite{seo2023lfs} performs best, probably because of utilizing a very strong baseline model StyleGAN2 \cite{karras2020analyzing}.  

\begin{table*}[t!]
\centering
\fontsize{4.7pt}{8pt}\selectfont
\caption{ Quantitative results comparison in terms of FID ($\downarrow$) and B--LPIPS ($\uparrow$) for each task with state-of-the-art methods.  }
\begin{tabular}{|l|cc|cc|cc|cc|cc|ll|} \hline  

           & \multicolumn{2}{|c|}{\textbf{Sketches ($T_1$)}} & \multicolumn{2}{|c|}{\textbf{Female ($T_2$)}} & \multicolumn{2}{|c|}{\textbf{Sunglasses ($T_3$)}} & \multicolumn{2}{|c|}{\textbf{Male ($T_4$)}} & \multicolumn{2}{|c|}{\textbf{Babies ($T_5$)}} & \multicolumn{2}{c|}{\textbf{Average}}\\  
           & FID   & B-LPIPS & FID   & B-LPIPS & FID   & B-LPIPS & FID   & B-LPIPS & FID   & B-LPIPS  &FID&B-LPIPS \\ \hline 
 CDC \cite{ojha2021few}& 322.72& 0.205& 197.40& 0.427& 244.94& 0.463& 277.00& 0.381& \underline{69.98}& 0.454& 208.41&0.386\\
 RSSA \cite{xiao2022few}& 308.00& 0.285& 175.20& 0.440& 207.58& 0.484& 205.49& 0.405& 76.70& 0.481& 194.59&0.419\\
 DCL \cite{zhao2022closer}& 297.73& 0.307& 170.31& 0.435& 191.54& 0.490& 194.42& 0.443& 77.22& 0.487& 186.25&0.432\\ \hline  
CAM-GAN \cite{cam_gan}& 91.81 & 0.293   & 85.68 & 0.332   & 86.81 & 0.333   & 82.83 & 0.312   & 146.20& 0.181     &98.66&0.290\\
LFS-GAN \cite{seo2023lfs}& \textbf{34.66}& \underline{0.354}& \textbf{29.59}&  \underline{0.481}& \textbf{27.69}& \underline{0.584}& \textbf{35.44}& \underline{0.472}& \textbf{ 41.48}& \underline{0.556}& \textbf{33.77}&\underline{0.489}\\ \hline  
CFTS-GAN (ours) & \underline{82.49} & \textbf{0.399}& \underline{62.10}& \textbf{0.707}& \underline{36.03}& \textbf{0.966}& \underline{66.23}&  \textbf{0.760}& 96.62& \textbf{1.02}&\underline{68.69}&\textbf{0.770}\\ \hline 
\end{tabular}
\label{fid-blpips}
\end{table*}

%

\subsection{Ablation Study}
To show the effectiveness of the CFTS-GAN we analyzed each component of our proposed method. The details for each component are given below.

\noindent\textbf{Effect of CDC on the teacher.} We analyzed the effect of CDC loss using different values of $w_t$ which are shown in Table \ref{ablation_teacher}. From the ablation, it is concluded that it increases diversity as we give more value to it. However, increasing $w_t$ after some point leads to a higher FID score. So, in our experiments, we use the optimal value of $w_t=40$ for training the teacher model for all of the tasks. It is because of gaining better FID at this point.

\begin{table}[h!]
\centering
\caption{Ablation study for the teacher using $w_t$ for sunglasses dataset.}
\begin{tabular}{|l|l|l|l|l|l|l|l|}
\hline
\multicolumn{1}{|c|}{$w_t$} & \multicolumn{1}{c|}{10} & \multicolumn{1}{c|}{20} & \multicolumn{1}{c|}{30} & \multicolumn{1}{c|}{40} & \multicolumn{1}{c|}{50} & \multicolumn{1}{c|}{60} & \multicolumn{1}{c|}{70} \\ \hline
\textbf{FID}                     & 79.41                   & 63.70                   & 46.94                   & \textbf{46.76}          & 54.21                   & 63.39                   & 65.76                   \\
\textbf{B-LPIPS}                 & 0.245                   & 0.516                   & 0.856                   & 0.930                   & 0.943                   & 0.956                   & \textbf{0.976}          \\ \hline
\end{tabular}
\label{ablation_teacher}
\end{table}

\noindent
\textbf{Effect of CDC on the student.} We also inspect the CDC loss terms and see its effect on the student model. Giving more weight to the CDC loss obtain more diversity and less weight leads to less diversity.  However, increasing its effect leads to degrading the FID as we analyzed for the teacher. We performed some ablation studies on the sunglasses dataset for different values of $w_s$ which is shown in Table \ref{ablation_student}. Assigning the $w_s=20$ works better for gaining more diversity with better FID. 
So, for all the tasks, we used $w_s=20$ for the student model.

\noindent 
\textbf{Effect of  $\mathcal{L}_{kd}$ on the student.} The experiments are also performed by assigning different values to $\alpha$.  We observe that giving the value of 2 leads to better results as shown in Table \ref{ablation_student}. Giving more value to it has more impact on diversity. Assigning greater value to it leads to more diversity and light improvement in the FID.  From this ablation study, we found when $\alpha =2$, we have better FID. Therefore, we use the mentioned value for the rest of the tasks. 

\begin{table*}[h!]
\centering
\fontsize{6pt}{9pt}\selectfont
\caption{Ablation study for the student on $\alpha$ and $w_s$ for the sunglasses dataset.}
\begin{tabular}{|l|llll|llll|llll|llll|}
\hline
                 & \multicolumn{4}{c|}{$\alpha = 0$}                                                                     & \multicolumn{4}{c|}{$\alpha = 2$}                                                                     & \multicolumn{4}{c|}{$\alpha = 5$}                                                                     & \multicolumn{4}{c|}{$\alpha = 10$}                                                                     \\ \hline
{\small $w_s$}& \multicolumn{1}{c|}{10} & \multicolumn{1}{c|}{20} & \multicolumn{1}{c|}{30} & \multicolumn{1}{c|}{40} & \multicolumn{1}{c|}{10} & \multicolumn{1}{c|}{20} & \multicolumn{1}{c|}{30} & \multicolumn{1}{c|}{40} & \multicolumn{1}{c|}{10} & \multicolumn{1}{c|}{20} & \multicolumn{1}{c|}{30} & \multicolumn{1}{c|}{40} & \multicolumn{1}{c|}{10} & \multicolumn{1}{c|}{20} & \multicolumn{1}{c|}{30} & \multicolumn{1}{c|}{40} \\ \hline
\textbf{FID}     & 56.35                   & 39.92                   & 40.98                   & 53.6                    & 50.66                   & \textbf{37.54}& 42.07                   & 54.31                   & 50.08                   & 38.55                   & 39.67                   & 47.87                   & 46.59                   & 42.80                   & 39.72                   & 46.94                   \\
\textbf{B-LPIPS} &                         0.588&                         0.942&                         0.990&                         1.01&                         0.768&                         0.944&                         \textbf{1.02}&                         1.01&                          0.851&                         0.993&                         0.991&                         1.00&                         0.917&                         0.997&                         1.00&                          1.01\\ \hline
\end{tabular}
\label{ablation_student}
\end{table*}

\noindent\textbf{Freezing student discriminator.}
We also analyzed the student model by freezing the layers of the discriminator of our CFTS-GAN. We inspect that training the last 24 layers of the discriminator leads to better FID scores and more diverse image generation instead of training all the layers (total layers are 36) of the discriminator.  The ablation study for the student model with a frozen discriminator is shown in Table \ref{ablation_freezD}.  For this ablation study, we consider the teacher with the best value as given in Table \ref{ablation_teacher}. From this analysis, it is concluded that instead of training all the layers, if some portion of the discriminator is kept frozen it leads to better results. 

\begin{table}[h!]
\centering
\caption{Ablation study for the student with freezing discriminator for sunglasses dataset.}
\begin{tabular}{|l|llllll|}
\hline
\multicolumn{1}{|c|}{Number of last trained layers} & \multicolumn{1}{c|}{6} & \multicolumn{1}{c|}{12} & \multicolumn{1}{c|}{18} & \multicolumn{1}{c|}{24} & \multicolumn{1}{c|}{30} & \multicolumn{1}{c|}{36}  \\ \hline
\textbf{FID}                     & 51.22     &  39.20                         & 38.98& \textbf{36.03}& 40.38& 37.54\\
\textbf{B-LPIPS}                 & 0.902                              & 0.910& 0.922& 0.966& \textbf{0.971}& 0.944\\ \hline
\end{tabular}
\label{ablation_freezD}
\end{table}

\section{Conclusion}\label{conclusion}
This article proposes a continual learning few shots generative adversarial network CFTS-GAN. The CFTS-GAN considers the challenging tasks of catastrophic forgetting and overfitting problems in GANs. We used the teacher-student model for the challenging task of continual few shots image generation. For continual learning, the CFTS-GAN uses adapter modules as a student for learning the new task while preserving the previously learned knowledge. The teacher model helps the student to produce better and more diverse image generation. The CDC is used by both the teacher and student to preserve the source diversity and prevent mode collapse.  Moreover,  we used a simple effective strategy of freezing the discriminator for more improvements.  To show the performance of the CFTS-GAN model, it is analyzed on different datasets. 
Which shows better results and produces diverse images than the state-of-the-art models.    


%
%

\bibliographystyle{splncs04}
\bibliography{bibliography}

%
\end{document}